\documentclass[11pt,a4paper]{article}
\usepackage[hyperref]{acl2021}
\usepackage{times}
\usepackage{latexsym}
\usepackage{graphicx}

\usepackage{microtype}

\aclfinalcopy 
\def\aclpaperid{157} 

\title{WVOQ at SemEval-2021 Task 6: BART for Span Detection and Classification}
\author{Cees Roele \\
	cees.roele@gmail.com}

\begin{document}
\maketitle
\begin{abstract}
\addcontentsline{toc}{section}{Abstract}
Simultaneous span detection and classification is a task not currently addressed in standard NLP frameworks. The present paper describes why and how an EncoderDecoder model was used to combine span detection and classification to address subtask 2 of SemEval-2021 Task 6. \par
\end{abstract}

\section{Introduction}
\addcontentsline{toc}{section}{Introduction}

Task 6 of SemEval-2021 studies the detection of persuasion techniques \cite{SemEval2021-6-Dimitrov}.  The task considers English language memes, which in subtasks are to be classified, divided into classified fragments having a begin and end, and classified when text is combined with images.\par

Of the three subtasks described in the paper, the present paper primarily addresses resolving subtask 2:\par

\begin{quotation}
Given only the "textual content" of a meme, identify which of the 20 techniques are used in it together with the span(s) of text covered by each technique. This is a multilabel sequence tagging task.
\end{quotation}

The figure below illustrates span detection and classification for three technique classes for a meme. \par

\begin{figure}[h!]
	\includegraphics[width=\linewidth]{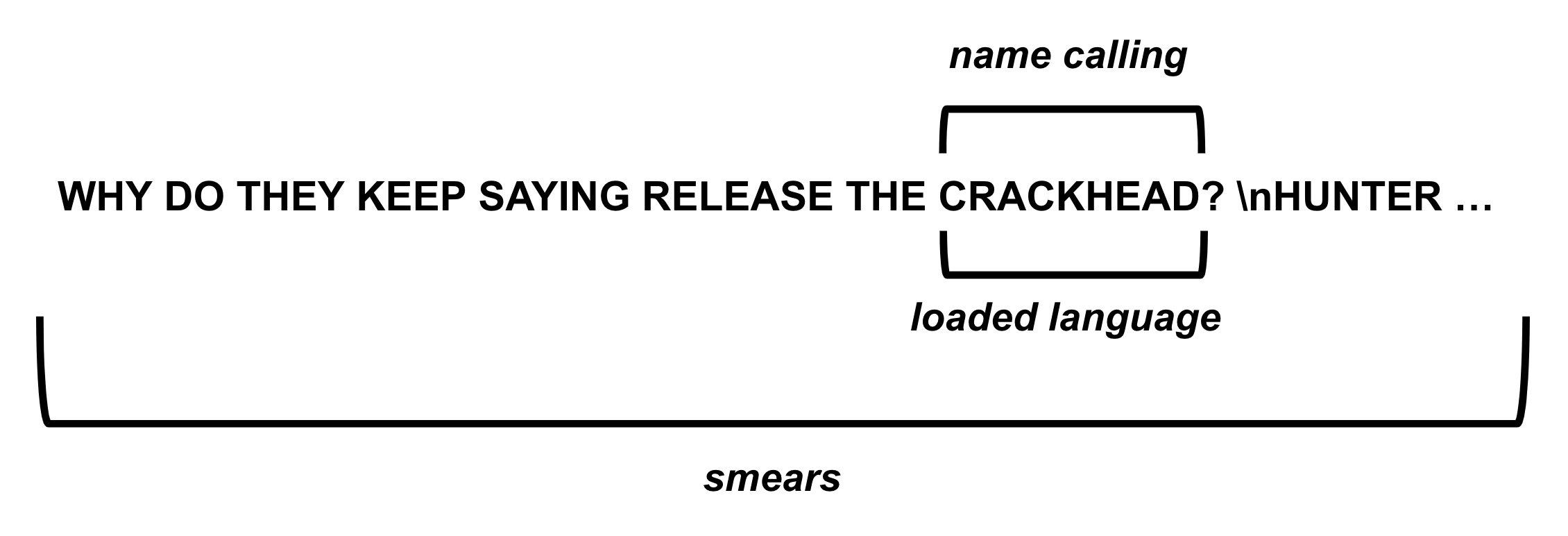}
	\caption{Span detection and classification: overlapping spans and spans extending over multiple sentences}
\end{figure}

In the above figure the ellipsis at the end of the sentence denotes the continuation of the second sentence.  Note  that \textit{loaded language} and \textit{name calling} both apply to the same span, that is, the word "CRACKHEAD".  Note furthermore, that the span for \textit{smears} overlaps with both of these spans and ranges over more than one sentence.\par

The present paper describes a novel approach to resolving these requirements by generating XML-like start and end tokens to delineate spans. The following illustrates this for the message in figure 1.\par

\small\texttt{\\$<$SMEARS$>$} \\
WHY DO THEY KEEP SAYING RELEASE THE \\
\texttt{$<$LOADED$-$LANGUAGE$>$ \\ 
$<$NAME$-$CALLING$>$} \\
CRACKHEAD \\
\texttt{$<$/NAME$-$CALLING$>$ \\
$<$/LOADED$-$LANGUAGE$>$} \\
? \symbol{92}n HUNTER ... \\
\texttt{$<$/SMEARS$>$}\\
\par
\normalsize

It attained an F1 score on the test set that is about in the middle of the baseline and the highest ranking score. \par

The choice of this approach of generating markup to identify spans was made on the basis that it was technically possible, easily understandable at a behavioral level of input and output,  and using a model that is pre-trained for dealing with spans. The aim was not so much to attain the highest score as to explore how effective this approach is in a proof-of-concept and what problems need to be overcome to bring it to good performance.\footnote{ The code for the described system is available at: \url{https://github.com/ceesroele/SemEval-2021-Task-6}.}\par

\section{Background}
\addcontentsline{toc}{section}{Background}

Propaganda messages are constructed using specific rhetorical techniques.  The current task is to identify within a message in what fragment a particular technique is invoked. \par

A simpler task is to identify fragments of a message in which any propaganda technique is used. Effectively, this comes down to classifying any part of a message as being either propaganda or not.  It is a sequence labeling problem that can be resolved for example using a BIO tagging format, where BIO stands for Begin, Inside, and Outside.  To classify a span of tokens as propaganda we can use \texttt{B-PROP} to designate the begin of the span, \texttt{I-PROP} to indicate the token being inside the earlier begun span, and \texttt{O} to designate a token not being part of a span.  \cite{chernyavskiy-etal-2020-aschern}. \par

For our case this approach can be extended by adding new labels for each technique, e.g. \texttt{B-SMEARS},  \texttt{I-SMEARS}, \texttt{B-NAME-CALLING}, \texttt{B-LOADED-LANGUAGE,} and so on for all twenty technique classes.  But looking at figure 1 we see that if CRACKHEAD is a token, we have to simultaneously label it as \texttt{I-SMEARS}, \texttt{B-NAME-CALLING}, and \texttt{B-LOADED-LANGUAGE}. The extension of the approach by just adding labels is not applicable to our situation in which spans can overlap.\par 

One solution for this problem is to retain the assumption that each input token is to be tagged, but add virtual depth.  This approach was taken for the \texttt{PRopaganda persuasion Techniques Analyzer} (\texttt{PRta}) \cite{da-san-martino-etal-2020-prta}. It is based on an architecture where each input token maps to as many output tokens as there are technique classes, plus one extra for \textit{no technique}.  Additionally, it uses a complementary output indicating confidence of any propaganda technique being present at the sentence level, which is used as a gate for predicting the presence of any specific techniques.\par

The sequence labeling method described at the beginning of this section is effectively a sequence-to-sequence translation, where the input and output sequence consist of the same number of tokens.  This allows us to match input with output based on position. To generate a marked up version of a message we need to allow an output sequence to have a length that differs from the input sequence.\par

By using an EncoderDecoder model we can generate arbitrary transformations of an input message including changing its length.  This can be used for abstractive dialogue, question answering,  and summarization.  A state of the art EncoderDecoder model is BART,  a denoising autoencoder built with a sequence-to-sequence model. \cite{lewis-etal-2020-bart}.  BART uses a
standard Tranformer-based neural machine translation architecture to couple a bidirectional encoder with a left-to-right decoder.  Pre-training BART was done by first corrupting text with an arbitrary noising function and then training a sequence-to-sequence model to reconstruct the original text. \par

\subsection{Data}
\addcontentsline{toc}{subsection}{Data}
There are two datasets available for task 6. The first is the Propaganda Techniques Corpus (PTC) dataset from SemEval-2020 Task 11. It consists of about 550 English language news articles in which spans - defined by begin and end positions - have been annotated with one out of 18 propaganda techniques. In practice a number of these techniques have been combined. For example, the three techniques \textit{whataboutism}, \textit{straw men}, and \textit{red herring} have been conflated into the single label \textit{whataboutism,straw\_men,red\_herring}. As a result, the dataset has effectively been annotated with 14 labels. Moreover, these composite labels don't identify individual labels in the 2021 dataset, which makes them unsuitable for training. That leaves only 12 usable labels in the PTC.\par

The 2021 dataset consists of about 660 English language memes. These are short texts consisting of mostly short sentences and relatively many uppercase characters.  Here fragments have been identified by start and end indexes and are labeled with one of a total of 20 classes. The differently labeled fragments may overlap, that is, a certain span of text may belong to fragments belonging to different classes.\par

The table below shows the number of fragments per dataset, the average number of words per fragment, the number of fragments spreading over more than one sentence, and the relative number of uppercase characters in fragments ( upper / (upper+lower)).

\begin{table}[ht]
\centering 
\begin{tabular}{l c c c c} 
\hline \textbf{Dataset} & \textbf{Spans} & \textbf{Words} &  \textbf{$>$} 1 & \textbf{Upper} \\ \hline 
PTC 2020 & 5610 & 8.6  & 223 & 0.04 \\ 
Memes 2021 & 1497 & 7.6  & 224 & 0.53 \\ 
\hline 
Total set & 7107 & 8.4 & 447 & 0.14  \\
\hline 
\end{tabular}
\caption{Data} 
\label{table:data} 
\end{table}

Regarding the data, we make the following observations:\par

\begin{itemize}
	\item For 8 of the classes there is data only in the relatively small 2021 memes dataset, which with many short sentences and a lot of uppercase is structurally different from the PCT 2020 dataset
	\item For some classes as much as half the characters in their fragments are in uppercase
	\item The median number of words in a fragment significantly varies per class. E.g.  \textit{smears}, \textit{causal oversimplification} , and \textit{whataboutism}  have median numbers of words of respectively 16, 20, and 25, while \textit{name calling/labeling}, \textit{loaded language} , and \textit{repetition}  have median numbers of words of respectively 3, 2, and 1.
	\item The median number of sentences  by fragments is 1 and for a handful of classes 2. 
\end{itemize}

The above findings will inspire a number of choices specified in the Experimental Setup below.\par

\subsection{Pre-training is key}
\addcontentsline{toc}{subsection}{Pre-training is key}
The success of language models like BERT \cite{devlin2019bert} derives in great part from a division of labor and domain. In the first step, a model is trained on a large body of unmarked data. This results in a model that has many linguistic relations represented in its weights, but that by itself is of little use. In the second step, that resulting pre-trained model is fine-tuned with data from a specific domain.\par

Given the comparative smallness of the two datasets at our disposal, leveraging pre-training can be expected to greatly enhance the quality of predictions. \par

However, it is worth considering what the pre-training entails. Take BERT. It was trained in part on English Wikipedia articles. But now we are looking at memes full of uppercase characters, containing persuasion techniques that we hope are not used in Wikipedia. Said differently, the data the model was pre-trained on might not be representative for our domain.\par

More abstract, but no less important, is the method of pre-training. Is the used method of Masked Language Modeling (MLM) supporting our task? We are interested in spans of text, possibly running across multiple sentences. Besides next sentence prediction, BERT’s methodology primarily consists of replacing a percentage of individual tokens with a mask token. However effective this may be, it is not optimized for spans.\par

SpanBERT \cite{joshi-etal-2020-spanbert} is effectively BERT trained with a different masking method:\par

\begin{itemize}
\item mask contiguous random spans, rather than random tokens, and 
\item train the span boundary representations to predict the entire content of the masked span, without relying on the individual token representations within it
\end{itemize}

SpanBERT outperforms BERT substantially on span selection tasks such as question answering and coreference resolution.\par

The present paper concerns a specific implementation for span detection and classification. Understanding pre-training helps us understand both why the presented system has a certain success and what its limitations are.\par

\section{System overview}
\addcontentsline{toc}{section}{System overview}

\subsection{Generating markup}

As sketched in the Background section, the problem we need to resolve is how to simultaneously represent a span of text and one of a multitude of labels.  Our solution is to step away from attempts to map onto a classification structure and instead regenerate the original text, but now with XML-like markup to indicate the start, end, and class of each fragment.\par

We regenerate the input text using an EncoderDecoder model.  Popularly expressed, it reads a text, and then generates a sequence of words. In order to add our markup for fragments we need two help functions, let's call them encipher and decipher. The encipher function takes text plus metadata on fragments and converts this into a string with XML-like markup. We need this to create our training data. The decipher function takes a string including XML-like markup and extracts metadata in the form of start, end, and class from it. \par

For each of the labels, the names for our technique classes, we create a start tag and an end tag. In order to let the tokenizer treat them each as single tokens, we add all these tags as tokens to the tokenizer.\par

\subsection{Using the BART EncoderDecoder model}

In principle it is possible to implement Encoder and Decoder on the basis of taking a pre-trained model for each, e.g. RoBERTa for the Encoder and BERT for the Decoder. Finding from a single trial was that in such a setup training went very slow and outcome was dissatisfactory. Instead we selected BART \cite{lewis-etal-2020-bart} as an integrated EncoderDecoder model.\par

BART has a number of pre-training methods that are of interest in trying to understand its performance. Only the first one is part of the pre-training methodology of BERT.
\begin{itemize}
\item \textbf{token masking}, like BERT
\item \textbf{token deletion}, random tokens are deleted and the model must decide which positions are missing tokens
\item \textbf{text infilling},  a number of spans with varying lengths are sampled and replaced with a single mask token.  Note that this is different from SpanBERT pre-training where each token of the span is replaced with a mask token.
\item \textbf{sentence permutation}, sentences are shuffled in random order
\item \textbf{document rotation}, a token is randomly chosen and the document is rotated to start with that token
\end{itemize}

We will get back to this when we evaluate the result.\par

\subsection{Easy to generalize}

Recuperating, we initialize the model by adding start and end tags for each technique class to the tokenizer. We use encipher  to create marked up versions of input texts to train the model.  To obtain fragments for given inputs we must decipher generated marked up texts to extract meta-data.  Besides having markup, the generated text may be different from the input. Directly deriving span positions from the markup tags leads to errors when that happens. This is to some degree mitigated by using an algorithm that searches for the best place of the tag in the original input string.

The novelty of the described system is in using a standard EncoderDecoder model to generate markup. No special architectural changes were made, no domain dependencies were introduced, and only minor pre- and postprocessing is done. It is therefor easy to turn the system into a general purpose span detection and classification system.

\section{Experimental setup}
\addcontentsline{toc}{section}{Experimental setup}

\subsection{Data and Training}

The articles of the PTC 2020 dataset were reduced to smaller segments on the basis of fragments. 
\small\begin{verbatim}
for each fragment:
  take all covering sentences
  while another fragment overlaps ..
       .. with any sentence in the segment
     add  fragment and those sentences
\end{verbatim}
\normalsize

Any sentences remaining, that is, not covered by any fragment, were ignored. The memes of the 2021 dataset were not split.

Mixing the 2020 and 2021 datasets for a single training run led to worse results than having a staged training of first the PTC 2020 data as pre-training and then the 2021 memes data as fine-tuning.  For training the datasets were split train:dev:test as 70:20:10. Training was done with a batch size of 8 for 25 epochs.

\subsection{Framework}

The system uses the \texttt{Seq2SeqModel} of Simple Transformers\footnote{ See: \url{https://simpletransformers.ai/}. The used version is 0.60.6. To be able to add begin and end markers as tokens to the seq2seq model a modification was made. It can be found in the github repository for the system discussed here, referred to in the first footnote. }, a task-oriented framework built on top of Hugging Face Transformers\footnote{\ See: \url{https://huggingface.co/transformers/ }. The used version is 4.3.2 }. It uses the Hugging Face pre-trained BART model identified with model type $"$bart$"$  and model name $"$facebook/bart-base$"$ . This is a model consisting of 6 encoder and decoder layers, 16-heads, and 139M parameters.\par

\subsection{Configuration}

Where training was done mostly with default settings, text generation required improved settings. We want enough tokens in the output for the full input plus markup, we want a relatively low penalty on length, to compensate for the previous setting, we want a relatively high penalty on repetition, and we perform a beam search. Experimentally, we came to the following settings as being optimal:\par

\begin{table}[ht]
\centering 
\begin{tabular}{l r} 
\hline \textbf{Parameter} & \textbf{Value} \\ \hline
max\_length & 200  \\ 
length\_penalty & 0.4  \\ 
repetition\_penalty & 2.0  \\ 
do\_sample & True  \\ 
num\_beams & 3  \\
top\_p & 0.8  \\
\hline 
\end{tabular}
\caption{Seq2SeqModel configuration} 
\label{table:configuration} 
\end{table}

\section{Results}
\addcontentsline{toc}{section}{Results}

The system's F1 score of 0.268 on subtask 2 on the test set scores about in the middle between the baseline and the highest ranking score.\par

\begin{table}[ht]
\centering 
\begin{tabular}{l l c c c} 
\hline \textbf{Rank} & \textbf{Team} & \textbf{F1 score} & \textbf{Precision} & \textbf{Recall} \\ \hline
1 & Volta & .482 & .501 & .464  \\ 
5 & WVOQ & .268 & .243 & .299 \\
   & baseline & .010 & .034 &  .006 \\
\hline 
\end{tabular}
\caption{Subtask 2 scores on the test set} 
\label{table:scores} 
\end{table}

Looking at errors we made the following observations:
\begin{itemize}
	\item Beginning and end tags in the generated text regularly don't match. 
	\item Generated text contains changed words and even added words, which leads to faulty identifications of spans.
\end{itemize}

Why does this happen? First, beginning and end tags are introduced as new tokens in the relatively small datasets we fine-tune with. Transformer models have no notion of syntactic connection between them and standard BART has not been pre-trained to relate these tokens correctly.  Second,  through its pre-training methodology BART is geared towards relative "freedom" in filling in spans. That's what makes it suitable for summarization and question-answering.  But what we need for markup generation is almost verbatim regeneration of the input.

\section{Conclusion}
\addcontentsline{toc}{section}{Conclusion}
The described system for span detection and simultaneous classification offers a proof$-$of$-$concept for a novel approach to sequence tagging based on generating a version of a message with markup for labels.  Its F1 score on the leaderboard is in the middle between the baseline and the top score. \par 

Drawback of the approach is that two types of systemic errors are introduced: tags lacking a matching tag, and tokens generated that are not in the original message.  These are not resolved by fine-tuning the model and they cannot be addressed with the standard configuration parameters of message generation in the sequence-to-sequence model.\par

Future research should aim at resolving these systemic errors.  Matching tags could be addressed through changes in the decoder's generation algorithm. Having the tokens in the output be the same as those in the input could be improved by amending the loss function for fine-tuning training of the model. \par

Only when these two issues are resolved will further optimization of the approach be worth investing effort in. \par 

\bibliography{bibliography-file}
\bibliographystyle{acl_natbib}
\end{document}